# INVESTIGATING THE PIECE-WISE LINEARITY AND BENCHMARK RELATED TO KOCZY-HIROTA FUZZY LINEAR INTERPOLATION

[1]MAEN ALZUBI, [2]SZILVESZTER KOVÁCS,

[1, 2] Department of Information Technology, University of Miskolc, H-3515 Miskolc, Hungary

E-mail: [1]alzubi@iit.uni-miskolc.hu, [2]szkovacs@iit.uni-miskolc.hu

**ABSTRACT**

Fuzzy Rule Interpolation (FRI) reasoning methods have been introduced to address sparse fuzzy rule bases and reduce complexity. The first FRI method was the Koczy and Hirota (KH) proposed "Linear Interpolation". Besides, several conditions and criteria have been suggested for unifying the common requirements FRI methods have to satisfy. One of the most conditions is restricted the fuzzy set of the conclusion must preserve a Piece-Wise Linearity (PWL) if all antecedents and consequents of the fuzzy rules are preserving on PWL sets at α-cut levels. The KH FRI is one of FRI methods which cannot satisfy this condition. Therefore, the goal of this paper is to investigate equations and notations related to PWL property, which is aimed to highlight the problematic properties of the KH FRI method to prove its efficiency with PWL condition. In addition, this paper is focusing on constructing benchmark examples to be a baseline for testing other FRI methods against situations that are not satisfied with the linearity condition for KH FRI.

## 1. INTRODUCTION

Fuzzy Rule Interpolation (FRI) is of particular importance for fuzzy reasoning when there is lacking knowledge or sparse fuzzy rule bases. If a given observation has no overlap with antecedent fuzzy sets, no rule included in classical fuzzy inference (e.g. Mamdani [1] and Sugeno [2]), and accordingly, no result can be obtained. However, FRI techniques introduced to enhance the robustness of fuzzy inference system and to reduce the complexity of fuzzy systems by excluding those rules which can be approximated by their adjacent ones. Further, it heightens the applicability of fuzzy systems by allowing an inevitable to be conclusion generated even if the existing fuzzy rule base does not cover a given observation.

FRI was presented to provide a reasonable and meaningful conclusion in case spares fuzzy rule bases, the first idea of the concept fuzzy interpolation was introduced by Koczy and Hirota in [3]-[8], which describes all fuzzy sets by a set of α-cuts ($\alpha \in (0,1]$). Given α-cuts, the interpolated conclusion fuzzy set can be calculated from α-cuts of the observation and all fuzzy sets involved in the fuzzy rule-based. Many FRI techniques suggested during the past two decades where the KH FRI was the base for many of them. Besides, several conditions for fuzzy interpolation techniques were recommended in [9]-[11] as a step towards unifying the fuzzy interpolation techniques which will be used for classification and comparison. One of the most conditions restricted to the conclusion fuzzy set is to preserve a Piece-Wise Linearity (PWL) if all the fuzzy rules and observation fuzzy sets are satisfied with PWL at α-cut levels.

The most significant benefit of the KH FRI is its low computational complexity that guarantees reasoning speed required by real-time applications, which produces the output based on its α-cuts. Notwithstanding many advantages, in some antecedent fuzzy set configuration, the KH FRI conclusion suffers from preserving a PWL (for more details see [12], [13]). The preserving of PWL is a necessary property that reflects how good the interpolative reasoning method handles the points between two consecutive α-cuts. Accordingly, the study in [14]-[16] discuss the PWL property and gives some conditions for the fuzzy rules and observation fuzzy sets, where the PWL of the conclusion necessarily holds.

The goal of this paper is to highlight the problematic properties of the KH FRI method to prove its efficiency with PWL condition in order to construct benchmark examples. This benchmark is set up to be a baseline for testing other FRI methods against cases that the KH FRI is not satisfied with the linearity condition. All benchmark examples in this paper are constructed using notations and equations in [14]-[16], that implemented by MATLAB FRI Toolbox [17], [18], which provides an easy-to-use framework to represent the FRI methods conclusions correctly and to know the expected results.

The rest of the paper is organized as follows: Section (2) introduces the background of the KH FRI with basic definitions related to fuzzy interpolative reasoning concept. The main equations and notations of the PWL property to KH FRI present in section (3) and the reference notations of the PWL property introduce in section (4). Benchmark examples of the KH FRI is constructed and presented in section (5). The results of the benchmark are discussed in section (6). Experiment some of the FRI methods according to benchmark examples in section (7). Finally, section (8) is dedicated to the conclusion of the paper.

## 2. NOTATIONS AND BASIC DEFINITIONS OF FUZZY RULE INTERPOLATION

The idea of the fuzzy interpolative reasoning technique was started depending on the concept of α-cuts distance, which is created initially for sparse fuzzy rule bases and complexity reduction that based on the resolution and extension principles, in which decompose the problem into an infinite family of crisp issues corresponding to α-cuts of fuzzy rule bases and observation. The interpolation conclusion can be solved for every α-cuts independently, and it can deduce the fuzzy solution by combining these results into a fuzzy approximation (see Equation (4)).

The KH FRI introduced as the first method for FRI concept, which knows as "linear interpolation" of two fuzzy rules for the area between their antecedents. The conclusion of the KH FRI could be calculated directly throughout generating an approximated conclusion from the observation and fuzzy rules [3]-[8], if the observation is located between two rule bases as follows:

$$A_1 \prec A^* \prec A_2$$
$$and$$
$$B_1 \prec B_2$$

Most FRI methods require some constraints to be satisfied: all the fuzzy sets of fuzzy rules and observation must be convex and normal, or briefly a CNF set. Let us assume (A) is a fuzzy set; thus, (A) is called normal when Height(A) = max(x) ∈ U(μA(x)), and is convex if each of its α-cuts are connected. Thus, the membership functions (MF) of fuzzy rules and observation (e.g. trapezoidal and triangular) are restricted to be PWL because it will be much easier for calculation with such functions because it depends on α-cuts. Some definitions could be introduced to realize the interpolation concept as follows:

**Definition 1**: Denotes the fuzzy sets of fuzzy rule bases and observation must be normal and convex on the universe of discourse $X_i$ by $P(X_i)$. Then for $A_1, A_2 \in P(X_i)$, if $\forall_\alpha \in (0,1]$, $A_1 \prec A_2$ if:

$$inf(A_{1\alpha}) < inf(A_{2\alpha}), \; sup(A_{1\alpha}) < sup(A_{2\alpha}) \quad (1)$$

Definition 2 describe the Fundamental Equation of Rule Interpolation (FERI), which is based on the concept of fuzzy distance [5] to all α-cut levels as follows:

**Definition 2**: Let $A_1 \rightarrow B_1$, $A_2 \rightarrow B_2$ be disjoint fuzzy rules on the universe of discourse $X \times Y$, and $A_1$, $A_2$ and $B_1$, $B_2$, be fuzzy sets on X and Y, respectively. Assume that $A^*$ is the observation of the input universe X. If $A_1 \prec A^* \prec A_2$, then the KH FRI between $R_1$ and $R_2$ are defined as follows:

$$d(A_1, A^*):d(A^*,A_2) = d(B_1, B^*):d(B^*,B_2) \quad (2)$$

where d refers to the fuzzy distance between fuzzy rule bases and observation fuzzy sets.

The conclusion $B^*$ of the KH FRI method can be generated directly based on α-cuts between fuzzy rules and observation fuzzy sets, which based on the upper (dU) and the lower (dL) fuzzy distances, where the similarity between the conclusion and the consequent must be the same between observation and antecedents. It can be calculated as follows:

**Definition 3**: Given a fuzzy relation $R\prec$: $(A_1, A_2)$ | $A_1, A_2 \in P(X)$, $A_1 \prec A_2$, if fuzzy sets $A_1$ and $A_2$ satisfy $R\prec$, the lower (dL) and the upper (dU) fuzzy distances between $A_1$ and $A_2$ by using the resolution principles [19], [20] as follows:

$$dL(A_1, A_2): R \prec \to P([0,1])$$
$$\mu dL(\delta): \sum \alpha \in [0,1] \; \alpha/d(\inf(A_{1\alpha}), \inf(A_{2\alpha}))$$
$$dU(A_1, A_2): R \prec \to P([0,1])$$
$$\mu dU(\delta): \sum \alpha \in [01] \; \alpha/d(\sup(A_{1\alpha}), \sup(A_{2\alpha}))$$

where $\delta \in [0,1]$ and d refers is the Euclidean distance or more generally, Minkowski distance.

**Definition 4**: Let $A_1$ and $A_2$ be fuzzy sets on the universe of discourse X with $|X| < \infty$, then the lower and upper distances between α-cuts sets $A_{1\alpha}$ and $A_{2\alpha}$ are defined as:

$$dL(A_{1\alpha}, A_{2\alpha}) = d(\inf(A_{1\alpha}), \inf(A_{2\alpha})),$$
$$dU(A_{1\alpha}, A_{2\alpha}) = d(\sup(A_{1\alpha}), \sup(A_{2\alpha})) \quad (3)$$

According to Definitions 3 and 4 the FERI of (dU) and (dL) α-cuts, the formula can be rewritten as follows:

$$dL(A^*, A_{1\alpha}) : dL(A^*, A_{2\alpha}) = dL(B^*, B_{1\alpha}) : dL(B^*, B_{2\alpha})$$
$$dU(A^*, A_{1\alpha}) : dU(A^*, A_{2\alpha}) = dU(B^*, B_{1\alpha}) : dU(B^*, B_{2\alpha})$$

Thus, the infimum (inf) and supremum (sup) of the conclusion can be determined:

$$\text{Inf}(B_\alpha^*) = \frac{dL(A_\alpha^*, A_{1\alpha}) \times \inf(B_{2\alpha}) + dL(A_\alpha^*, A_{2\alpha}) \times \inf(B_{1\alpha})}{dL(A_\alpha^*, A_{1\alpha}) + dL(A_\alpha^*, A_{2\alpha})}$$

$$\text{Sup}(B_\alpha^*) = \frac{dU(A_\alpha^*, A_{1\alpha}) \times \sup(B_{2\alpha}) + dU(A_\alpha^*, A_{2\alpha}) \times \sup(B_{1\alpha})}{dU(A_\alpha^*, A_{1\alpha}) + dU(A_\alpha^*, A_{2\alpha})}$$

Then,

$$B_\alpha^* = (\inf(B_\alpha^*); \sup(B_\alpha^*)).$$

Finally, the consequence $B^*$ can be constructed by Equation (4):

$$B^* = \bigcup_{\alpha \in [0,1]} \alpha \cdot B_\alpha^* \quad (4)$$

Figure 1 represents the linear interpolation method between two fuzzy rule bases and observation described by trapezoidal membership function (MF) for $\alpha \in [0, 1]$. The characteristic points of the trapezoidal MF denoted by vector a= $[a_1, a_2, a_3, a_4]$, where the support ($a_1$ and $a_4$) represents by P(0, L) and P(0, U), the core ($a_2$ and $a_3$) describes by P(1, L) and P(1, U), in which L denotes to lower, and U denotes to upper. In case of triangular MF, it can be represented by P(0, L), P(0, U) and P(1, (L and U)) where $a_2 = a_3$ for the core fuzzy set (A).

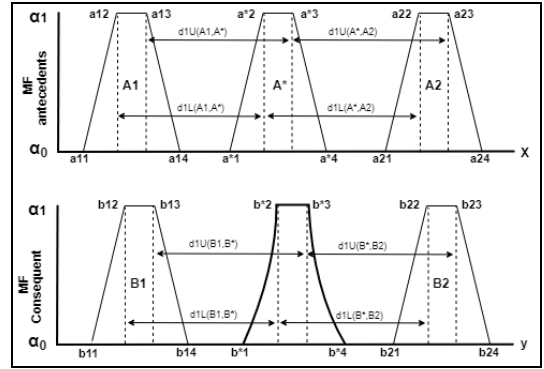

*Figure 1: The Ratio of the Lower and Upper Distances Calculated Between the Interpolation of Two Piece-Wise Linear Rules. The Shape of the Conclusion ($B^*$) Shows for the α -Cuts Level Between α $\in$ (0, 1) [14]*

## 3. A PIECE-WISE-LINEARITY OF THE KH FRI BASED ON α -CUT LEVELS

Most of the FRI techniques proposed are based on linear interpolation, e.g. KH FRI. In which the conditions and criteria proposed for unifying the requirements of the FRI methods have to satisfy. Therefore, some of the necessary restrictions of the fuzzy linear interpolation methods required all the fuzzy sets of fuzzy rules and observation must be CNF sets. Furthermore, the fuzzy sets are also restricted to preserve PWL. Hence, most FRI methods do not preserve PWL in conclusion (see cases in [21]). The KH FRI is the one, which cannot fulfil this condition and failed the demand for a PWL conclusion.

The convexity property of the conclusion fuzzy set can be checked if all α-cuts are connected. Hence, it will be checked that the FRI methods are preserving the PWL for α $\in$ [0, 1]. Contrarily, interpolation techniques implemented will not produce any results if α-cuts are not connected since they are represented as intervals.

Most applications are restricted to a small finite set of α-cut levels, which will be called the necessary cuts. For PWL membership functions (e.g. trapezoidal and triangular), an obvious assumption is to define the set of significant cuts by the united breakpoint set α. However, this is not true in general because of most cases for interpolation methods, $B^*$ is severely distorted and non-linear, if the α $\in$ [0, 1] levels are unknown.

Theoretically, the conclusion of the KH FRI can be calculated by its α-cuts, where all α-cuts should be considered, but for practical reasons, only a finite set is taken into consideration during the computation. Now, let us determine which notations will be used to calculate the characteristic points of the lower and upper of fuzzy rule bases and observation fuzzy sets that can be defined as follows:

For antecedent fuzzy set:

$$A_{i\alpha L} = \alpha \cdot (a_{i2} - a_{i1}) + a_{i1}$$
$$A_{i\alpha U} = \alpha \cdot (a_{i3} - a_{i4}) + a_{i4}$$

For consequent fuzzy set:

$$B_{i\alpha L} = \alpha \cdot (b_{i2} - b_{i1}) + b_{i1}$$
$$B_{i\alpha U} = \alpha \cdot (b_{i3} - b_{i4}) + b_{i4}$$

For observation fuzzy set:

$$A^*_{\alpha L} = \alpha \cdot (a^*_2 - a^*_1) + a^*_1$$
$$A^*_{\alpha U} = \alpha \cdot (a^*_3 - a^*_4) + a^*_4$$

Furthermore, the conclusion of the linear interpolation (left slope) could be calculated for α-cut levels by the statement as follows:

**Statement 1**: The equations of the left and right slopes to breakpoint levels 0 and α can be calculated for the two fuzzy rule bases $A_1 \rightarrow B_1$, $A_2 \rightarrow B_2$ and the observation $A^*$ as follows:

$$B^*_{\alpha L} = \frac{DL_1 \times \alpha^2 + DL_2 \times \alpha + DL_3}{cL_9 \times \alpha + cL_{10}} \quad (5)$$

$$B^*_{\alpha R} = \frac{DR_1 \times \alpha^2 + DR_2 \times \alpha + DR_3}{cR_9 \times \alpha + cR_{10}} \quad (6)$$

where

$DL_1 = (cL_3.cL_5)+(cL_1.cL_7)$
$DL_2 = (cL_3.cL_6)+(cL_4.cL_5)+(cL_1.cL_8)+(cL_2.cL_7)$
$DL_3 = (cL_4.cL_6)+(cL_2.cL_8)$

And

$cL_1 = a^*_2 - a^*_1 - a_{12} + a_{11}$; $cL_2 = a^*_1 - a_{11}$
$cL_3 = a_{22} - a_{21} - a^*_2 + a^*_1$; $cL_4 = a_{21} - a^*_1$
$cL_5 = b_{12} - b_{11}$; $cL_6 = b_{11}$
$cL_7 = b_{22} - b_{21}$; $cL_8 = b_{21}$
$cL_9 = a_{11} - a_{12} + a_{22} - a_{21}$;
$cL_{10} = a_{21} - a_{11}$

Similar to the left slope equation, the right slope can be constructed, it can replace the index (1) of the characteristic points fuzzy set $a_{1(1)}$, $a_{2(1)}$, $a^*_{(1)}$, $b_{1(1)}$ and $b_{2(1)}$ by index (4), and index (2) of $a_{1(2)}$, $a_{2(2)}$, $a^*_{(2)}$, $b_{1(2)}$ and $b_{2(2)}$ are replaced by (3), and the sign in X replaced by its opposite (negative direction tangents).

On the other hand, authors in [14]-[16] also introduced other equations to calculate the left and right slopes of the conclusion as follows:

The left slope of the conclusion:

$$B^*_L = \frac{(DL_3.cL_9^2) - (DL_2.cL_9.cL_{10}^2) + (DL_1.cL_{10}^2)}{cL_9^2.(cL_9.\alpha + cL_{10})} \times \frac{DL_1}{cL_9}.\alpha + \frac{(DL_2.cL_9) - (DL_1.cL_{10})}{cL_9^2} \quad (7)$$

it can be written:

$$\frac{A}{\alpha + B} + (C.\alpha + D) = yH + yL$$

where

$$A = \frac{(DL_3.cL_9^2) - (DL_2.cL_9.cL_{10}^2) + (DL_1.cL_{10}^2)}{cL_9^3}$$

$$B = \frac{cL_{10}}{cL_9}, \quad C = \frac{DL_1}{cL_9}, \quad D = \frac{(DL_2.cL_9) - (DL_1.cL_{10})}{cL_9^2}$$

where the $y_L$ refer to a straight line and $y_H$ denotes the hyperbola, $B_{\alpha L}$ is the curve that represent the superposition of $y_L$ and $y_H$ (for more details see Figure 3 in [14], [16]). The right slope can be calculated by similar equations of the left slop.

## 4. REFERENCE VALUES FOR THE PWL PROPERTY

Regarding the conclusion of the KH FRI method is not fulfilled on preserving a PWL, i.e. in general, the fundamental equation applied between two adjacent fuzzy rule bases and observation for the α levels is not linear, it slightly deviates from the calculated linear interpolation. According to the main corollaries in [14]-[16], the linearity of the left and right slopes to the KH FRI conclusion could be determined as follows:

The condition of polynomiality is very simple when ($cL_9 = 0$). Then, we get:

**Corollary 1:** The flanks of $B^*$ are a piece-wise polynomial if and only if the two antecedents $A_1$ and $A_2$ have equivalent PWL slopes, obtainable from each other by geometric translations:

$$a_{12} - a_{11} = a_{22} - a_{21}$$

As well, if we require linearity of the pieces, the condition must be met, when ($DL_1 = 0$). Consequently, the linearity conclusion can be demonstrated:

**Corollary 2:** This corollary will be satisfied in three different cases that slopes of the conclusion ($B^*$) are preserving a PWL. Hence, if this corollary is done suitably, the KH FRI conclusion will always be satisfied if the following cases are held:

**Case C1.1:** If the left and right slopes of the antecedents $A_i$ and the consequents $B_i$ are equivalent to PWL on the universe of discourse.

The left slope notations can be defined as:

$$A_i = a_{12} - a_{11} = a_{22} - a_{21}$$
$$B_i = b_{12} - b_{11} = b_{22} - b_{21}$$

**Case C1.2:** If the left and right slopes and characteristic points of the two adjacent fuzzy rule bases $A_1 \Rightarrow B_1$ and $A_2 \Rightarrow B_2$ are equivalent on the universe of discourse.

The left slope notations can be determined as follows:

$$A_1 \Rightarrow B_1: a_{12} - a_{11} = b_{12} - b_{11}$$
$$A_2 \Rightarrow B_2: a_{22} - a_{21} = b_{22} - b_{21}$$

In this case, there is no restriction on the shape of the observation $A^*$.

**Case C2:** If the antecedents $A_i$ and the observation $A^*$ are satisfied with PWL. The $B^*$ slopes are linear only if Corollary 1 is applied.

The left slope notations can be determined as follows:

$$d = d^*$$

where

$$a_{22} - a_{21} = a_{21} - a_{11} = d$$
$$a^*_2 - a^*_1 = d^*$$

For this case, there is no restriction on the consequents $B_i$.

**Case C3:** If all the variables on the universe of discourse are covered by equidistant fuzzy sets $A_i$, $B_i$ and $A^*$.

Notations of the left slope can be described as follows:

$$A_i = a_{12} - a_{11} = a_{22} - a_{21}$$
$$B_i = b_{12} - b_{11} = b_{22} - b_{21}$$
$$A^* = a^*_2 - a^*_1$$

In [14]-[16], the upper bound is presented the possible highest deviation between the real and approximated linear functions, hence, if there is a large difference between them, the validity of the method is violated between characteristic points of the fuzzy sets based on the intervals [0, 1], and at the same time could question the applicability of any new method. Regarding the beneficial computational properties of the KH FRI would not hold any more. Consequently, different views were introduced to determine the deviation from the calculated linear interpolation. Therefore, the approximating linear equation of the conclusion defined to give a straight line that will be used to compare with real function. It can be determined as follows:

For the left slope of conclusion $B^*$ to two endpoints are:

$$B^*_{0L} = \frac{DL_3}{DL_{10}}, \quad B^*_{1L} = \frac{DL_1 + DL_2 + DL_3}{cL_9 + cL_{10}}$$

Then, the equation of the left slope of the linear approximation is determined as:

$$B^*_{\alpha L(approx)} = \alpha \times (B^*_{1L} - B^*_{0L}) + B^*_{0L} \qquad (8)$$

Figure 2 describes the maximum difference between the real function and its PWL approximation, which can be determined by statement 2:

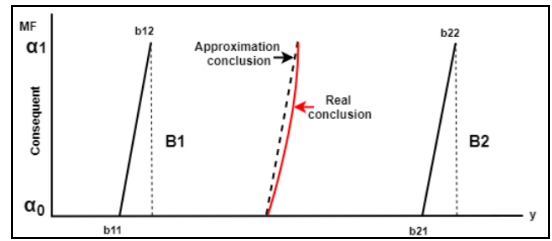

*Figure 2: The Difference Between the Linear Approximation and Real Function of the Left Slope for the α-Cuts Level Between $\alpha \in (0, 1)$ [14], [16]*

**Statement 2:** The error of approximating the nonlinear slope of the determined conclusion by a linear slope between (0 and 1) expressed in terms of the membership degree running through [0, 1]:

$$\Delta B_{\alpha L}^* = \frac{DL_1}{cL_9 \times \alpha + cL_{10}} \times \alpha^2 + \left( \frac{DL_2}{cL_9 \times \alpha + cL_{10}} + \frac{DL_3}{cL_{10}} - \frac{DL_1 + DL_2 + DL_3}{cL_9 + cL_{10}} \right) \times \alpha + \left( \frac{DL_3}{(cL_9 \times \alpha) + cL_{10}} - \frac{DL_3}{cL_{10}} \right) \quad (9)$$

The Equation in (7) could be used to verify the PWL condition. Further notations presented to check the upper limit of the error can be given by calculating the difference $y_H(0) - y_H(1)$ (for more details see [14], [16]), which can be determined as follows:

$$E = yH(0) - yH(1) = \frac{A}{B(1+B)}$$

$$\frac{(DL_3.cL_9^2) - (DL_2.cL_9.cL_{10}) + (DL_1.cL_{10}^2)}{cL_9.cL_{10}.(cL_9 + cL_{10})} \quad (10)$$

Consequently, the linearity error can be determined as:

**Statement 3:** The linearity error of $B^*_L$ (for the left slope) does not exceed $\varepsilon > 0$ if:

$$\frac{(DL_2 + cL_{10}.\varepsilon) + \sqrt{(DL_2 + cL_{10}.\varepsilon)^2 - 4.DL_1.(DL_3 - cL_{10}.\varepsilon)}}{2.(DL_3 - cL_{10}.\varepsilon)}$$

$$\leq \frac{cL_9}{cL_{10}} = \frac{1}{B} \quad (11)$$

$$\frac{(DL_2 + cL_{10}.\varepsilon) - \sqrt{(DL_2 + cL_{10}.\varepsilon)^2 - 4.DL_1.(DL_3 - cL_{10}.\varepsilon)}}{2.(DL_3 - cL_{10}.\varepsilon)}$$

which can be proved by:

For the left slope:

$$Left.slope = ((DL_3 - (cL_{10} \times \varepsilon)) \times cL_9^2) - ((DL_2 \times cL_{10} + cL_{10}^2 \times \varepsilon) \times cL_9 + (DL_1 \times cL_{10}^2)) \leq 0$$

For the right slope:

$$Right.slope = ((DR_3 - (cR_{10} \times \varepsilon)) \times cR_9^2) - ((DR_2 \times cR_{10} + cR_{10}^2 \times \varepsilon) \times cR_9 + (DR_1 \times cR_{10}^2)) \leq 0$$

where the value $\varepsilon$ is assumed 0 to verify notations of the statement 3.

The general case of the linear interpolation can only use two breakpoint values ($\alpha = 0$ and $\alpha = 1$) for computing the support and the core of the conclusion, which may not be satisfactory because in most cases the results obtained are somewhat disappointing. For this reason, it will be needed to calculate for a much larger number of α-cuts levels. In the next section, we will discuss all cases that will be used in constructing the benchmark examples. These cases will be analyzed according to PWL condition, which values of α-cut levels to every step of 0.1, $\alpha \in [0, 1]$ will be considered.

## 5. THE PWL BENCHMARK OF THE KH FRI

In this section, benchmark examples will be constructed to demonstrate the validate of the PWL condition of the KH FRI method, in which statements and equations in the previous section could be used to check the linearity conclusion of the KH FRI method, and also to construct the benchmark examples. The left and right slopes of the fuzzy rule bases and observation play a significant role in preserving the linearity conclusion.

The benchmark examples constructed using one-dimensional of input and output variables, the triangular membership function and two fuzzy rules are used to represent fuzzy sets of the antecedent, consequent and observation. All benchmark examples and its results tested by MATLAB FRI toolbox. The current version of FRI toolbox is freely available to download in [17].

Benchmark examples are divided into two groups. The first group presents the conclusions of the KH FRI method are satisfied with PWL condition. The second group shows the conclusions of the KH FRI are not satisfied with PWL condition. Now, we will discuss in details the cases of the KH FRI conclusion to PWL.

The KH FRI conclusion is always satisfied with PWL condition if the following cases are met:

**For Case C1.1:** When the left and right slopes $A_i$ and $B_i$ fuzzy sets are identical (e.g. for left slop $a_{12} - a_{11} = a_{22} - a_{21}$ and $b_{12} - b_{11} = b_{22} - b_{21}$). The conclusion of KH FRI will always be satisfied with the linearity condition. Table 1 illustrates notations of the Example X1 that demonstrate the linearity conclusion related to Case C1.1.

*Table 1: The Preserving PWL Conclusion of the KH FRI with Fuzzy Sets and Notations to Case C1.1.*

| Example X1 | | |
|---|---|---|
| **The characteristic points of the fuzzy sets:** $A_1$=[0 2 2 6] $A_2$=[10 12 12 16] $A^*$=[7 8 8 9] $B_1$=[0 2 2 6] $B_2$=[10 12 12 16] $B^*$=[7 8 8 9] | | **The length of left and right slopes of the fuzzy sets:** **For left**: $A_1$=2, $A_2$=2, $A^*$=1, $B_1$=2, $B_2$=2 **For Right**: $A_1$=4, $A_2$=4, $A^*$=1, $B_1$=4, $B_2$=4 |
| **By notations in (9):** $\Delta B^*$ Left = 0 $\Delta B^*$ Right = 0 | **By notations in (10):** E.Left = NAN E.Right = NAN | **By notations in (11):** Left.Slope = 1 Right.Slope = 1 |

**For Case C1.2:** If two adjacent fuzzy rule bases $A_1 \rightarrow B_1$ and $A_2 \rightarrow B_2$ (e.g. for left slop: Rule1 ($a_{12} - a_{11} = b_{12} - b_{11}$), Rule2 ($a_{22} - a_{21} = b_{22} - b_{21}$) have the same left and right slopes and the same characteristic points on the universe of discourse. Then, the KH FRI conclusion will always be satisfied with the linearity condition. Table 2 explains the Example X2 that indicate to Case C1.2.

*Table 2: The Preserving PWL Conclusion of the KH FRI with Fuzzy Sets and Notations to Case C1.2.*

| Example X2 | | |
|---|---|---|
| **The characteristic points of the fuzzy sets:** $A_1$=[0 3 3 4] $A_2$=[10 11 11 14] $A^*$=[5 6 6 7] $B_1$=[0 3 3 4] $B_2$=[10 11 11 14] $B^*$=[5 6 6 7] | | **The length of left and right slopes of the fuzzy sets:** **For left**: $A_1$=3, $A_2$=1, $A^*$=1, $B_1$=3, $B_2$=1 **For Right**: $A_1$=1, $A_2$=3, $A^*$=1, $B_1$=1, $B_2$=3 |
| **By notations in (9):** $\Delta B^*$ Left = 0 $\Delta B^*$ Right = 0 | **By notations in (10):** E.Left = 0 E.Right = 0 | **By notations in (11):** Left.Slope = 1 Right.Slope = 1 |

**For Case C2:** When the fuzzy sets of the antecedents $A_i$ and the observation $A^*$ have the same left and right slopes PWL. Then, the conclusion of the KH FRI will always be satisfied with the linearity condition. Table 3 defined notations of Example X3 regard to Case C2.

*Table 3: The Preserving PWL Conclusion of the KH FRI with Fuzzy Sets and Notations to Case C2.*

| Example X3 | | |
|---|---|---|
| **The characteristic points of the fuzzy sets:** $A_1$=[0 3 3 6] $A_2$=[13 16 16 19] $A^*$=[6.5 9.5 9.5 12.5] $B_1$=[1 2 2 3] $B_2$=[7 9 9 11] $B^*$=[4 5.5 5.5 7] | | **The length of left and right slopes of the fuzzy sets:** **For left**: $A_1$=3, $A_2$=3, $A^*$=3, $B_1$=1, $B_2$=2 **For Right**: $A_1$=3, $A_2$=3, $A^*$=3, $B_1$=1, $B_2$=2 |
| **By notations in (9):** $\Delta B^*$ Left = 0 $\Delta B^*$ Right = 0 | **By notations in (10):** E.Left = NAN E.Right = NAN | **By notations in (11):** Left.Slope = 1 Right.Slope = 1 |

**For Case C3:** When the left and right slopes for all fuzzy sets of two adjacent fuzzy rule bases and observation are equidistant ($A_i = B_i = A^*$). Therefore, the conclusion of the KH FRI will always be satisfied with the linearity condition. Table 4 illustrates notations for Example X4 which indicate to Case C3.

*Table 4: The Preserving PWL Conclusion of the KH FRI with Fuzzy Sets and Notations to Case C3.*

| Example X4 | | |
|---|---|---|
| **The characteristic points of the fuzzy sets:** $A_1$=[1 2 2 3] $A_2$=[10 11 11 12] $A^*$=[5 6 6 7] $B_1$=[1 2 2 3] $B_2$=[10 11 11 12] $B^*$=[5 6 6 7] | | **The length of left and right slopes of the fuzzy sets:** **For left**: $A_1$=1, $A_2$=1, $A^*$=1, $B_1$=1, $B_2$=1 **For Right**: $A_1$=1, $A_2$=1, $A^*$=1, $B_1$=1, $B_2$=1 |
| **By notations in (9):** $\Delta B^*$ Left = 0 $\Delta B^*$ Right = 0 | **By notations in (10):** E.Left = NAN E.Right = NAN | **By notations in (11):** Left.Slope = 1 Right.Slope = 1 |

However, the conclusions of the KH FRI are not satisfied with PWL condition based on Equations (9), (10) and (11) if the following cases are held.

**According to Case C1.1:** When the left and right slopes $A_i$ and $B_i$ are incompatible (e.g. for left slop ($a_{12} - a_{11} \neq a_{22} - a_{21}$) and ($b_{12} - b_{11} = b_{22} - b_{21}$) whereas $A_i \neq A^*$, in this case, the linearity conclusion of KH FRI is not satisfied. Example Y1 constructed to prove the problem, which will be described by three different situations based on the characteristic points of the observation $A^*$ to compare its linearity conclusions. Table 5 illustrates notations that describe the problem according to the three situations.

*Table 5: The Problem with Slopes to Case C1.1 Which Is Not Preserving PWL*

| Example Y1 situation 1 when e.g. the left slope (b1(2) - b1(1) = b2(2) - b2(1)) = $A^*$ | | |
|---|---|---|
| **The characteristic points of the fuzzy sets:** $A_1$=[0 2 2 8] $A_2$=[14 20 20 22] $A^*$=[9 11 11 13] $B_1$=[0 2 2 4] $B_2$=[9 11 11 13] $B^*$=[5.79 6.50 6.50 7.21] | | **The length of left and right slopes of the fuzzy sets:** **For left**: $A_1$=2, $A_2$=6, $A^*$=2, $B_1$=2, $B_2$=2 **For Right**: $A_1$=6, $A_2$=2, $A^*$=2, $B_1$=2, $B_2$=2 |
| **By notations in (9):** $\Delta B^*$ Left (Maximum-deviation) = 0.08 $\Delta B^*$ Right (Maximum-deviation) = 0.08 | **By notations in (10):** E.Left = 1.2857 E.Right = 1.2857 | **By notations in (11):** Left.Slope = 0 Right.Slope = 0 |
| Example Y1 situation 2 when e.g. the left slope (b1(2) - b1(1) = b2(2) - b2(1)) < $A^*$ | | |
| **The characteristic points of the fuzzy sets:** $A_1$=[0 2 2 8] $A_2$=[14 20 20 22] $A^*$=[8 11 11 14] $B_1$=[0 2 2 4] $B_2$=[9 11 11 13] $B^*$=[5.14 6.50 6.50 7.86] | | **The length of left and right slopes of the fuzzy sets:** **For left**: $A_1$=2, $A_2$=6, $A^*$=3, $B_1$=2, $B_2$=2 **For Right**: $A_1$=6, $A_2$=2, $A^*$=3, $B_1$=2, $B_2$=2 |
| **By notations in (9):** $\Delta B^*$ Left (Maximum-deviation) = 0.04 $\Delta B^*$ Right (Maximum-deviation) = 0.04 | **By notations in (10):** E.Left = 0.6429 E.Right = 0.6429 | **By notations in (11):** Left.Slope = 0 Right.Slope = 0 |
| Example Y1 situation 3 when e.g. the left slope (b1(2) - b1(1) = b2(2) - b2(1)) > $A^*$ | | |

### Table 5 (continued)

| Example Y1 (continued) |  |
|---|---|
| The characteristic points of the fuzzy sets: $A_1$=[0 2 2 8] $A_2$=[14 20 20 22] $A^*$=[10 11 11 12] $B_1$=[0 2 2 4] $B_2$=[9 11 11 13] $B^*$=[6.4286 6.5000 6.5000 6.571] | The length of left and right slopes of the fuzzy sets: **For left**: $A_1$=2, $A_2$=6, $A^*$=1, $B_1$=2, $B_2$=2 **For Right**: $A_1$=6, $A_2$=2, $A^*$=1, $B_1$=2, $B_2$=2 |
| By notations in (9): $\Delta B^*$ Left (Maximum-deviation) = 0.121 $\Delta B^*$ Right (Maximum-deviation) = 0.121 | By notations in (10): E.Left = 1.9286 E.Right = 1.9286 | By notations in (11): Left.Slope = 0 Right.Slope = 0 |

**About Case C1.2:** When the two adjacent fuzzy rule bases $A_1 \rightarrow B_1$ and $A_2 \rightarrow B_2$ have the same left and right slopes but have different characteristic points on the universe of discourse, in this case, the linearity conclusion of KH FRI is not satisfied. Example Y2 constructed to prove the issue as shown on Table 6.

*Table 6: The Problem with Slopes to Case C1.2 Which Is Not Preserving PWL*

| Example Y2 |  |
|---|---|
| The characteristic points of the fuzzy sets: $A_1$=[0 3 3 4] $A_2$=[10 11 11 14] $A^*$=[5 6 6 7] $B_1$=[1 4 4 5] $B_2$=[15 16 16 19] $B^*$=[8 8.5 8.5 9.2] | The length of left and right slopes of the fuzzy sets: **For left**: $A_1$=3, $A_2$=1, $A^*$=1, $B_1$=3, $B_2$=1 **For Right**: $A_1$=1, $A_2$=3, $A^*$=1, $B_1$=1, $B_2$=3 |
| By notations in (9): $\Delta B^*$ Left (Maximum-deviation) = 0.028 $\Delta B^*$ Right (Maximum-deviation) = 0.017 | By notations in (10): E.Left = 0.500 E.Right = 0.300 | By notations in (11): Left.Slope = 0 Right.Slope = 0 |

**Referring to Case C2:** When the left and right slopes of the antecedents $A_i$ ($a_{12} - a_{11} = a_{22} - a_{21}$) and the observation $A^*$ are not equivalent whereas $A_i \neq B_i$, then, the linearity conclusion of KH FRI is not satisfied. Refer to corollary 1, Example Y3 is applied the polynomial condition when ($a_{12} - a_{11} = a_{22} - a_{21}$), however, is not linear, Table 7 describes notations which prove the problem to this case.

*Table 7: The Problem with Slopes to Case C2 Which Is Not Preserving PWL*

| Example Y3 |  |
|---|---|
| The characteristic points of the fuzzy sets: $A_1$=[0 3 3 7] $A_2$=[15 18 18 22] $A^*$=[7 8 8 10] $B_1$=[0 2 2 5] $B_2$=[8 9 9 10] $B^*$=[3.7333 4.3333 4.3333 6.0000] | The length of left and right slopes of the fuzzy sets: **For left**: $A_1$=3, $A_2$=3, $A^*$=1, $B_1$=2, $B_2$=1 **For Right**: $A_1$=4, $A_2$=4, $A^*$=2, $B_1$=3, $B_2$=1 |
| By notations in (9): $\Delta B^*$ Left (Maximum-deviation) = 0.033 $\Delta B^*$ Right (Maximum-deviation) = 0.067 | By notations in (10): E.Left = NAN E.Right = NAN | By notations in (11): Left.Slope = 0 Right.Slope = 0 |

**According to Case C3:** When values of the left and right slopes of fuzzy rule bases and observation are not similar ($A_i \neq B_i \neq A^*$), in this case, the linearity conclusion of KH FRI is not satisfied. Example Y4 created to demonstrate the problem as shown on Table 8.

*Table 8: The Problem with Slopes to Case C3 Which Is Not Preserving PWL*

| Example Y4 |  |
|---|---|
| The characteristic points of the fuzzy sets: $A_1$=[1 2 2 4] $A_2$=[10 12 12 15] $A^*$=[6 7 7 8] $B_1$=[0 2 2 5] $B_2$=[12 13 13 14] $B^*$=[6.6667 7.5 7.5 8.2727] | The length of left and right slopes of the fuzzy sets: **For left**: $A_1$=1, $A_2$=2, $A^*$=1, $B_1$=2, $B_2$=1 **For Right**: $A_1$=2, $A_2$=3, $A^*$=1, $B_1$=3, $B_2$=1 |
| By notations in (9): $\Delta B^*$ Left (Maximum-deviation) = 0.031 $\Delta B^*$ Right (Maximum-deviation) = 0.101 | By notations in (10): E.Left = 1.1667 E.Right = 4.2273 | By notations in (11): Left.Slope = 0 Right.Slope = 0 |

## 6. DISCUSSION OF THE PWL BENCHMARK EXAMPLES

In this section, the benchmark examples and its notations that have been created will be discussed to demonstrate the linearity of the KH FRI conclusions. Examples X1 to X4 that shown on Tables 1-4 proved the conclusions of the KH FRI are always satisfied with PWL, which were determined by Equation (9) that is always equal to 0 because the values of the real and linear approximation functions are similar. Also, by Equation (10) that is NAN or in some cases is equal to 0 because the parameters $cL_9$ or $cR_9$ are equal to 0 (see corollary 1 and 2). The Equation (11) could always be satisfied with preserving the PWL when left.Slope and right.Slope are 1.

According to the examples of the first group, two examples will be taken to prove the linearity of the KH FRI.

Referring to Example X1 on Table 1, the conclusion of the KH FRI is satisfied with PWL condition related to Case C1.1, where the support length of the left and right slopes of antecedent $A_i$ and consequent $B_i$ fuzzy sets are similar, e.g. the left slope of antecedent fuzzy sets: ($A_1$=3 and $A_2$=3) and ($B_1$=2 and $B_2$=2). Therefore, Equations (9), (10) and (11) are always satisfied with the linearity conclusion. Figure 3 describes the result of $\Delta B^*$ by Equation (9) for all α-cut levels to the left and right slopes that are equal to 0. On the other side, the estimated error by Equation (10) is NAN for left and right slopes (E.left = NAN, E.right = NAN), the notations of corollary 1 and 2 are also demonstrated, where (e.g. for left slope) $DL_1$ is 0 (because of $cL_9$ = 0).

| | Example X1 | | | | | | | | | | |
|---|---|---|---|---|---|---|---|---|---|---|---|
| α - Levels | | 0.000 | 0.100 | 0.200 | 0.300 | 0.400 | 0.500 | 0.600 | 0.700 | 0.800 | 0.900 | 1.000 |
| Left Slope | Real B* | 5.000 | 5.100 | 5.200 | 5.300 | 5.400 | 5.500 | 5.600 | 5.700 | 5.800 | 5.900 | 6.000 |
| | approx B* | 5.000 | 5.100 | 5.200 | 5.300 | 5.400 | 5.500 | 5.600 | 5.700 | 5.800 | 5.900 | 6.000 |
| | ΔB* | 0.000 | 0.000 | 0.000 | 0.000 | 0.000 | 0.000 | 0.000 | 0.000 | 0.000 | 0.000 | 0.000 |
| Right Slope | Real B* | 6.000 | 6.100 | 6.200 | 6.300 | 6.400 | 6.500 | 6.600 | 6.700 | 6.800 | 6.900 | 7.000 |
| | approx B* | 6.000 | 6.100 | 6.200 | 6.300 | 6.400 | 6.500 | 6.600 | 6.700 | 6.800 | 6.900 | 7.000 |
| | ΔB* | 0.000 | 0.000 | 0.000 | 0.000 | 0.000 | 0.000 | 0.000 | 0.000 | 0.000 | 0.000 | 0.000 |

*Figure 3: The Difference Between the Linear Approximation and Real Functions of the Left and Right Slopes for α ∈ [0,1] to Example X1*

Another case of preserving linearity, it is evident by Example X2 on Table 2, the conclusion of the KH FRI is satisfied with linearity condition when slopes of two fuzzy rule bases are equivalent, e.g. Rule 1 (for the lower $A_1=3$ and $B_1=3$), and (for the upper $A_1=1$ and $B_1=1$). Therefore, ΔB* of the left and right slopes that are equal 0 which computed by Equation (9). Also, the estimated error according to Equation (10) is 0 (E.left = 0, E.right = 0), despite the parameters $cL_9$ and $DL_1$ (by notations of the Equation (6)) are "not zero", e.g. for left slope, $cL_9 = -2$ and $DL_1$ are -2 because this example is restricted to the characteristic points of the two fuzzy rule bases that must be identical as mentioned in Case C1.2.

In contrast, Examples Y1-Y4 describe cases where the conclusions of the KH FRI are not satisfied with PWL (the second group). These examples have been presented based on two facts, either if the conclusion is close to linearity (Example Y1 situation 2) or far from linearity (Example Y1 situation 3). According to Equations (9) and (10) will be discussed in details for each example. Regarding the Equation (11) will be not satisfied with PWL condition because the value of this equation is always 0 for both (left.Slope) and (right.Slope).

According to Example Y1 (Case C1.1) on Table 5, it illustrates three different situations based on the characteristic points of the observation as: situation 1 (when $(b_{12} − b_{11} = b_{22} − b_{21}) = A^*$), situation 2 (when $(b_{12} − b_{11} = b_{22} − b_{21}) < A^*$) and situation 3 (when $(b_{12} − b_{11} = b_{22} − b_{21}) > A^*$). Figure 4 explains the difference between real and linear approximation functions for each one. According to Equation (9), the maximum deviation for left and right slopes in situation 1 is 0.08, and situation 2 is smaller than situation 1 which is 0.04, in contrast, situation 3 has the high deviation is 0.121. On the other hand, the Equation (10) describes the error ratios, which are different for three situations, situation 3 has a large error ratio compared to situation 1 and situation 2, where the error ratio of the left and right slopes of situation 3 is 1.9286, situation 1 is 1.2857, and situation 2 is 0.6429. Then, situation 3 is far away from linearity, in contrast to situation 2 which is closer than situation 1 to linearity.

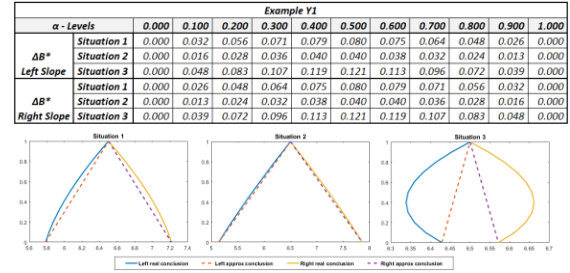

| | Example Y1 | | | | | | | | | | |
|---|---|---|---|---|---|---|---|---|---|---|---|
| α - Levels | | 0.000 | 0.100 | 0.200 | 0.300 | 0.400 | 0.500 | 0.600 | 0.700 | 0.800 | 0.900 | 1.000 |
| ΔB* Left Slope | Situation 1 | 0.000 | 0.032 | 0.056 | 0.071 | 0.079 | 0.080 | 0.075 | 0.064 | 0.048 | 0.026 | 0.000 |
| | Situation 2 | 0.000 | 0.016 | 0.028 | 0.036 | 0.040 | 0.040 | 0.038 | 0.032 | 0.024 | 0.013 | 0.000 |
| | Situation 3 | 0.000 | 0.048 | 0.083 | 0.107 | 0.119 | 0.121 | 0.113 | 0.096 | 0.072 | 0.039 | 0.000 |
| ΔB* Right Slope | Situation 1 | 0.000 | 0.026 | 0.048 | 0.064 | 0.075 | 0.080 | 0.079 | 0.071 | 0.056 | 0.032 | 0.000 |
| | Situation 2 | 0.000 | 0.013 | 0.024 | 0.032 | 0.038 | 0.040 | 0.040 | 0.036 | 0.028 | 0.016 | 0.000 |
| | Situation 3 | 0.000 | 0.039 | 0.072 | 0.096 | 0.113 | 0.121 | 0.119 | 0.107 | 0.083 | 0.048 | 0.000 |

*Figure 4: The Difference Between the Linear Approximation and Real Functions of the Left and Right Slopes for α ∈ [0,1] to Example Y1*

In Example Y2 on Table 6 which illustrates the problem when the left and right slopes of fuzzy rule bases are the same, but the characteristic points of the fuzzy sets of $A_i$ and $B_i$ are different on the universe of discourse. In this case, the conclusion KH FRI is not satisfied with linearity. Referring to Equation (9), the deviation for the left slope is greater than the right slope, where the left slope is 0.028, and the right slope is 0.017, as shown in Figure 5. Also, the Equation (10) introduced the error ratio, where the left slope is 0.500 is far from linearity to the right slope is 0.300.

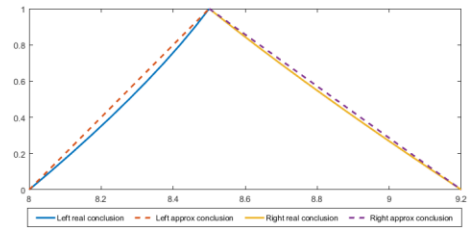

| Example Y2 | | |
|---|---|---|
| α - Levels | ΔB* | |
| | Left Slope | Right Slope |
| 0.000 | 0.000 | 0.000 |
| 0.100 | 0.009 | 0.007 |
| 0.200 | 0.017 | 0.011 |
| 0.300 | 0.022 | 0.015 |
| 0.400 | 0.026 | 0.016 |
| 0.500 | 0.028 | 0.017 |
| 0.600 | 0.027 | 0.016 |
| 0.700 | 0.024 | 0.013 |
| 0.800 | 0.019 | 0.010 |
| 0.900 | 0.011 | 0.006 |
| 1.000 | 0.000 | 0.000 |

*Figure 5: The Difference Between the Linear Approximation and Real Functions of the Left and Right Slopes for α ∈ [0,1] to Example Y2*

In Example Y3 on Table 7, it demonstrates the conclusion of KH FRI is not satisfied with linearity according to the results of equations and notations. Figure 6 explains the difference between real and linear approximation functions where the maximum deviation for left and right slopes are 0.033 and 0.067 respectively by (9). Equation (10) describes the error ratio for the left and the right slopes are NAN, by referring to corollary 1, this example has achieved the condition of polynomiality because the left and right slopes of $A_1$ and $A_2$ are similar, but not linear.

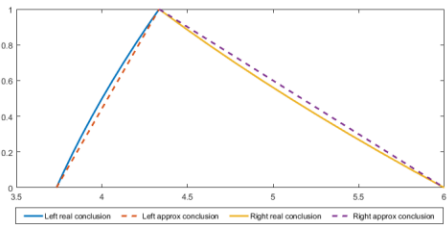

| Example Y3 | | |
|---|---|---|
| α - Levels | ΔB* | |
| | Left Slope | Right Slope |
| 0.000 | 0.000 | 0.000 |
| 0.100 | 0.012 | 0.024 |
| 0.200 | 0.021 | 0.043 |
| 0.300 | 0.028 | 0.056 |
| 0.400 | 0.032 | 0.064 |
| 0.500 | 0.033 | 0.067 |
| 0.600 | 0.032 | 0.064 |
| 0.700 | 0.028 | 0.056 |
| 0.800 | 0.021 | 0.043 |
| 0.900 | 0.012 | 0.024 |
| 1.000 | 0.000 | 0.000 |

*Figure 6: The Difference Between the Linear Approximation and Real Functions of the Left and Right Slopes for α ∈ [0,1] to Example Y3*

Also, in Example Y4 on Table 8, all fuzzy sets of fuzzy rule bases and observation are different, then, the conclusion KH FRI is also not satisfied with the linearity condition. The error ratio of the linearity in the right slope is 4.2273 which is so far than in left slope is 1.1667. Additionally, Figure 7 defines the difference between real and linear approximation functions as follows:

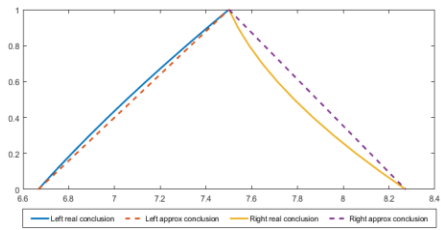

| Example Y4 | | |
|---|---|---|
| α - Levels | ΔB* | |
| | Left Slope | Right Slope |
| 0.000 | 0.000 | 0.000 |
| 0.100 | 0.409 | 0.207 |
| 0.200 | 0.535 | 0.365 |
| 0.300 | 0.556 | 0.475 |
| 0.400 | 0.525 | 0.538 |
| 0.500 | 0.467 | 0.556 |
| 0.600 | 0.390 | 0.529 |
| 0.700 | 0.303 | 0.459 |
| 0.800 | 0.207 | 0.347 |
| 0.900 | 0.106 | 0.193 |
| 1.000 | 0.000 | 0.000 |

*Figure 7: The Difference Between the Linear Approximation and Real Functions of the Left and Right Slopes for α ∈ [0,1] to Example Y4*

## 7. COMPARING SOME OF THE FRI METHODS BASED ON PWL BENCHMARK

In this section, the FRI methods (KHstab [12], VKK [13], FRIPOC [22] and VEIN [23]) will be compared by the constructed benchmark to the KH method. To offer a simple way of comparison we focused on the cases that the KH FRI method demonstrated the fails of preserving PWL, which was represented by Examples (Y1$_{situation\ 1,2,3}$, Y2, Y3 and Y4) as shown on Tables 5-8. Therefore, this comparison shows the difference between the results of the selected methods related to the PWL property for each example. A multi levels of α were are used to test these comparisons.

Figures 8, 9, 10 and 11 describe the results of the FRI methods (KHstab, VKK, FRIPOC and VEIN) and illustrate the difference between real conclusion (red line) and approximate conclusion (black line).

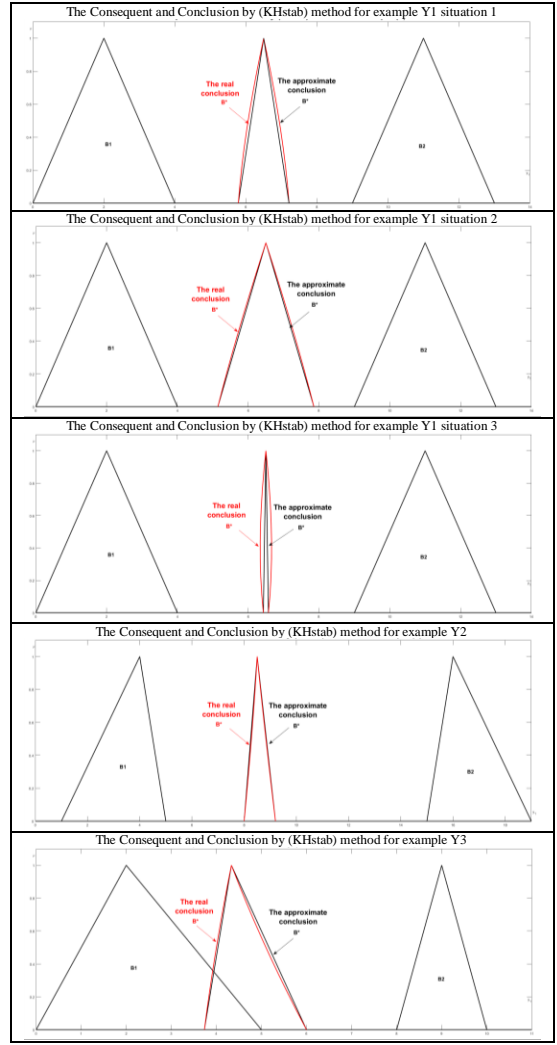

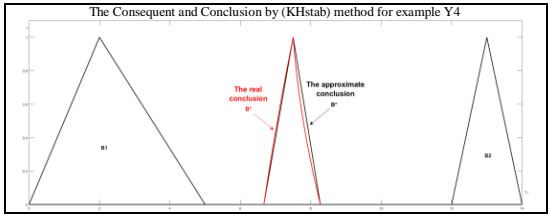

*Figure 8: The Approximation and Real Conclusions of the KHstab [12] Method to Examples ($Y1_{situation\ 1,2,3}$, Y2, Y3 and Y4).*

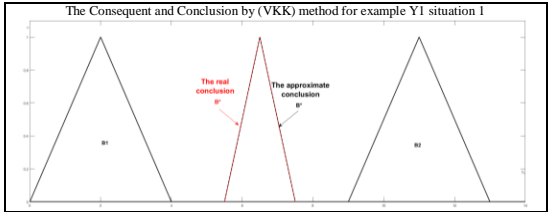
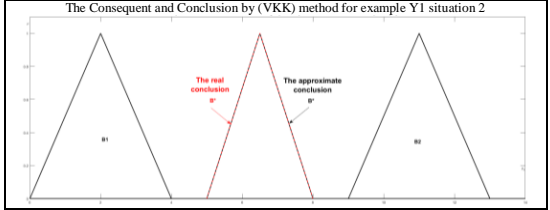
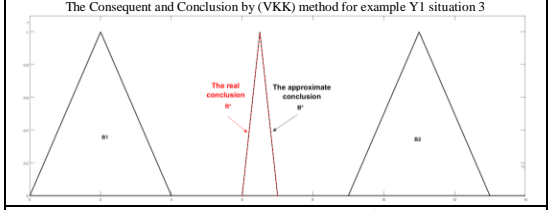
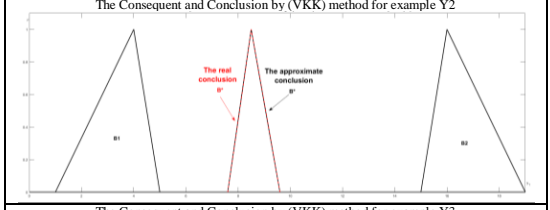
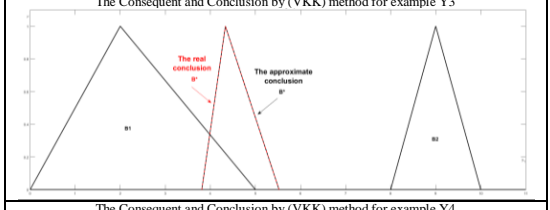
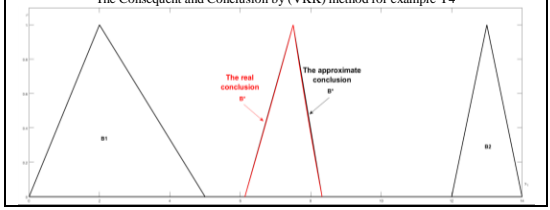

*Figure 9: The Approximation and Real Conclusions of the VKK [13] Method to Examples ($Y1_{situation\ 1,2,3}$, Y2, Y3 and Y4).*

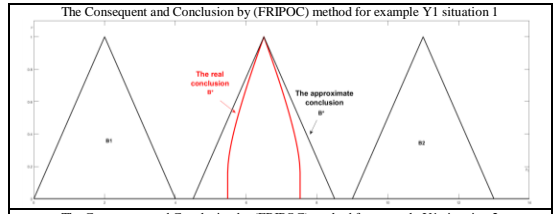
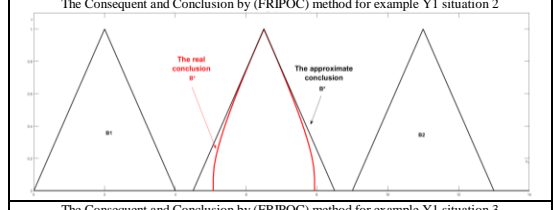
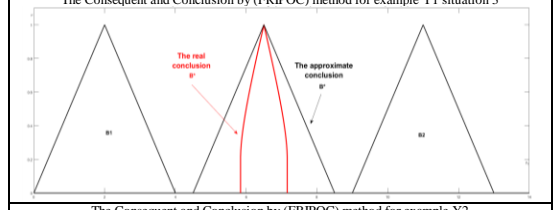
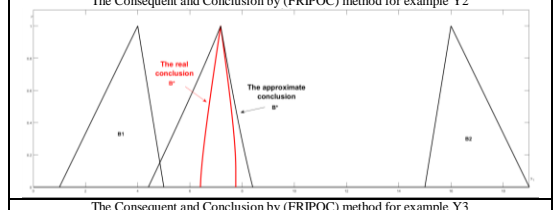
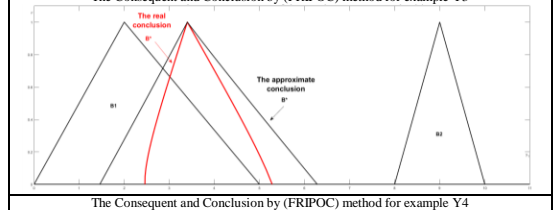
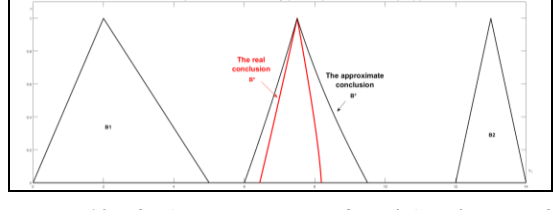

*Figure 10: The Approximation and Real Conclusions of the FRIPOC [22] Method to Examples ($Y1_{situation\ 1,2,3}$, Y2, Y3 and Y4).*

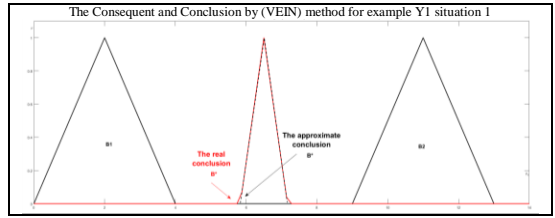

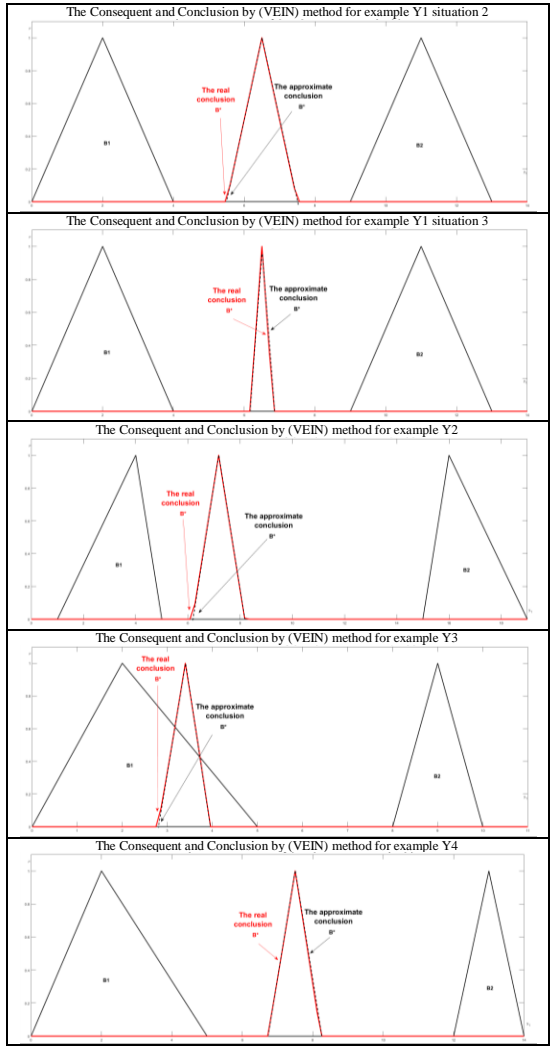

*Figure 11: The Approximation and Real Conclusions of the VEIN [23] Method to Examples (Y1$_{situation\ 1,2,3}$, Y2, Y3 and Y4)*

According to the results of the FRI methods (KHstab [12], VKK [13], FRIPOC [22] and VEIN [23]) to benchmark Examples (Y1$_{situation\ 1,2,3}$, Y2, Y3 and Y4) of the KH FRI method, we conclude the following:

- KHstab and FRIPOC methods are not fulfilled with preserving on PWL property to all benchmark Examples (Y1$_{situation\ 1,2,3}$, Y2, Y3 and Y4).
- VKK method succeeded with preserving on PWL property in all benchmark examples, except Example Y4, which has appeared with a little bit deviation in the right side.
- VEIN method succeeded with PWL property on benchmark Examples (Y1$_{situation\ 1,2}$, Y2 and Y3), in contrast, the Examples (Y1$_{situation\ 3}$ and Y4) have appeared with a little bit deviation in the bottom boundary.

Table 9 presents a summary for evaluation selected FRI methods according to benchmark Examples (Y1$_{situation\ 1,2,3}$, Y2, Y3 and Y4) to PWL property, where the plus sign (+) indicates the technique is satisfied with PWL property, while a minus sign (-) shows the method has a little bit deviation. The plus sign (x) indicates the technique did not preserve on PWL property.

*Table 9: Summary of the FRI methods and their conformity to the benchmark Examples (Y1$_{situation\ 1,2,3}$, Y2, Y3 and Y4).*

| Example | Methods | | | |
|---|---|---|---|---|
| | KHstab [12] | VKK [13] | FRIPOC [22] | VEIN [23] |
| Y1 situation1 | x | + | x | - |
| Y1 situation2 | x | + | x | - |
| Y1 situation3 | x | + | x | + |
| Y2 | x | + | x | - |
| Y3 | x | + | x | - |
| Y4 | x | - | x | + |

## 8. CONCLUSIONS

FRI techniques introduced as an alternative for classical inference system, many conditions and criteria were suggested as a step to unify FRI methods, but there are no particular examples to compare between the FRI methods, one of the most important conditions is preserving PWL, where the conclusions of the interpolation technique require to preserve PWL in case all fuzzy sets of fuzzy rule base are PWL. In this study, we determined the necessary and sufficient notations and equations that demonstrate the PWL property to the KH FRI method, which proposed as the first method of FRI concept. Also, we discussed the relationship between the linear approximation and real function conclusions for the left and right slopes and tested within several levels of α-cuts. Then, we constructed special benchmark examples to prove the KH FRI satisfies and fails the requirements for the PWL conclusion. This benchmark aimed to be used as a reference for evaluation and comparison with other FRI methods.

Finally, some of the FRI methods (KHstab, VKK, FRIPOC and VEIN) were compared based on PWL benchmark Examples (Y1$_{situation\ 1,2,3}$, Y2, Y3 and Y4) that are shown the KH FRI is not satisfied with PWL property. Hence, the results of the FRI methods as shown in Table 9 illustrated the KHstab and FRIPOC methods are suffered to the preservation of PWL property, in contrast, VKK and VEIN methods succeeded in preserving PWL property except for some benchmark examples with a little bit deviation in linearity.

## ACKNOWLEDGMENTS

The described study was carried out as part of the EFOP3.6.1-16-00011 Younger and Renewing University - Innovative Knowledge City - institutional development of the University of Miskolc aiming at intelligent specialization project implemented in the framework of the Szechenyi 2020 program. The realization of this project is supported by the European Union, co-financed by the European Social Fund.